\pdfoutput=1
\documentclass[11pt,a4paper]{article}
\usepackage[hyperref]{emnlp2020}
\usepackage{times}
\usepackage{latexsym}
\usepackage{graphicx, subfigure, multirow}
\usepackage{amsmath}
\usepackage{bbm}
\usepackage{amsfonts}

\newcommand{\argmin}{\mathop{\arg\min}}

\usepackage{microtype}

\aclfinalcopy 
\def\aclpaperid{3476} 

\title{Compressing Transformer-Based Semantic Parsing Models using Compositional Code Embeddings}

\author{Prafull Prakash$^{1} \thanks{~~Equal contribution, alphabetical order}$~, Saurabh Kumar Shashidhar$^{1}$\footnotemark[1]~, Wenlong Zhao$^{1}$\footnotemark[1]~, \\
{\bf Subendhu Rongali$^1$, Haidar Khan$^2$, \and Michael Kayser$^2$} \\
  $^1$University of Massachusetts Amherst \\
  $^2$Amazon Alexa \\
  \texttt{\{prafullpraka, ssaurabhkuma, wenlongzhao, srongali\}@cs.umass.edu}\\
  \texttt{\{khhaida, mikayser\}@amazon.com}}

\begin{document}
\maketitle
\begin{abstract}
The current state-of-the-art task-oriented semantic parsing models use BERT or RoBERTa as pretrained encoders; these models have huge memory footprints. This poses a challenge to their deployment for voice assistants such as Amazon Alexa and Google Assistant on edge devices with limited memory budgets. We propose to learn compositional code embeddings to greatly reduce the sizes of BERT-base and RoBERTa-base. We also apply the technique to DistilBERT, ALBERT-base, and ALBERT-large, three already compressed BERT variants which attain similar state-of-the-art performances on semantic parsing with much smaller model sizes. We observe 95.15\% $\sim$ 98.46\% embedding compression rates and 20.47\% $\sim$ 34.22\% encoder compression rates, while preserving $>$97.5\% semantic parsing performances. We provide the recipe for training and analyze the trade-off between code embedding sizes and downstream performances.
\end{abstract}

\section{Introduction}
Conversational virtual assistants, such as Amazon Alexa, Google Home, and Apple Siri, have become increasingly popular in recent times. These systems can process queries from users and perform tasks such as playing music and finding locations. A core component in these systems is a task-oriented semantic parsing model that maps natural language expressions to structured representations containing intents and slots that describe the task to perform. For example, the expression \emph{Can you play some songs by Coldplay?} may be converted to \emph{Intent: PlaySong, Artist: Coldplay}, and the expression \emph{Turn off the bedroom light} may be converted to \emph{Intent: TurnOffLight, Device: bedroom}. 

Task-oriented semantic parsing is traditionally approached as a joint intent classification and slot filling task. \citet{kamath2018survey} provide a comprehensive survey of models proposed to solve this task. Researchers have developed semantic parsers based on Recurrent Neural Networks \cite{mesnil2013investigation, liu2016attention, hakkani2016multi}, Convolutional Neural Networks \cite{xu2013convolutional, kim2014}, Recursive Neural Networks \cite{guo2014joint}, Capsule Networks \cite{sabour2017, zhang2019}, and slot-gated attention-based models \cite{goo2018slot}. 

The current state-of-the-art models on SNIPS \cite{coucke2018snips}, ATIS \cite{price1990evaluation}, and Facebook TOP \cite{gupta2018semantic} datasets are all based on BERT-style \cite{bert, liu2019roberta} encoders and transformer architectures \cite{chen2019bert,castellucci2019multi,rongali}. It is challenging to deploy these large models on edge devices and enable the voice assistants to operate locally instead of relying on central cloud services, due to the limited memory budgets on these devices. However, there has been a growing push towards the idea of TinyAI \footnote{https://www.technologyreview.com/technology/tiny-ai/}.

In this paper, we aim to build space-efficient task-oriented semantic parsing models that produce near state-of-the-art performances by compressing existing large models. We propose to learn compositional code embeddings to significantly compress BERT-base and RoBERTa-base encoders with little performance loss. We further use ALBERT-base/large \cite{albert} and DistilBERT \cite{distilbert} to establish light baselines that achieve similar state-of-the-art performances, and apply the same code embedding technique. We show that our technique is complementary to the compression techniques used in ALBERT and DistilBERT. With all variants, we achieve 95.15\% $\sim$ 98.46\% embedding compression rates and 20.47\% $\sim$ 34.22\% encoder compression rates, with $>$97.5\% semantic parsing performance preservation.

\section{Related Compression Techniques}
\subsection{BERT Compression}
Many techniques have been proposed to compress BERT \cite{bert}. \citet{ganesh2020compressing} provide a survey on these methods. Most existing methods focus on alternative architectures in transformer layers or learning strategies. 

In our work, we use DistilBERT and ALBERT-base as light pretrained language model encoders for semantic parsing. DistilBERT \cite{distilbert} uses distillation to pretrain a model that is 40\% smaller and 60\% faster than BERT-base, while retaining 97\% of its downstream performances. ALBERT \cite{albert} factorizes the embedding and shares parameters among the transformer layers in BERT and results in better scalability than BERT. ALBERT-xxlarge outperforms BERT-large on GLUE \cite{GLUEdataset}, RACE \cite{lai2017race}, and SQUAD \cite{rajpurkar2016squad} while using less parameters.

We use compositional code learning \cite{codebooks} to compress the model embeddings, which contain a substantial amount of model parameters. Previously ALBERT uses factorization to compress the embeddings. We find more compression possible with code embeddings.

\subsection{Embedding Compression}
Varied techniques have been proposed to learn compressed versions of non-contextualized word embeddings, such as, Word2Vec \cite{mikolov2013efficient} and GLoVE \cite{pennington2014glove}. \citet{spine} use denoising k-sparse autoencoders to achieve binary sparse intrepretable word embeddings. \citet{chen2016} achieve sparsity by representing the embeddings of uncommon words using sparse linear common combination of common words. \citet{word2bits} achieve compression by quantization of the word embeddings by using 1-2 bits per parameter. \citet{faruqui} use sparse coding in a dictionary learning setting to obtain sparse, non-negative word embeddings. \citet{raunak2017} achieve dense compression of word embeddings using PCA combined with a post-processing algorithm. \citet{codebooks} propose to represent word embeddings using compositional codes learnt directly in end-to-end fashion using neural networks. Essentially few common basis vectors are learnt and embeddings are reconstructed using their composition via a discrete code vector specific to each token embedding. This results in 98\% compression rate in sentiment analysis and 94\% - 99\% in machine translation tasks without performance loss with LSTM based models. All the above techniques are applied to embeddings such as WordVec and Glove, or LSTM models. 

We aim to learn space-efficient embeddings for transformer-based models. We focus on compositional code embeddings \cite{codebooks} since they maintain the vector dimensions, do not require special kernels for calculating in a sparse or quantized space, can be finetuned with transformer-based models end-to-end, and achieve extremely high compression rate. \citet{chen2018learning} explores similar idea as \citet{codebooks} and experiment with more complex composition functions and guidances for training the discrete codes. \citet{chen2019differentiable} further show that end-to-end training from scratch of models with code embeddings is possible. Given various pretrained language models, we find that the method proposed by \citet{codebooks} is straightforward and perform well in our semantic parsing experiments.

\section{Method}
\subsection{Compositional Code Embeddings}
\citet{codebooks} apply additive quantization \cite{babenko2014additive} to learn compositional code embeddings to reconstruct pretrained word embeddings such as GloVe \cite{pennington2014glove}, or task-specific model embeddings such as those from an LSTM neural machine translation model. Compositional code embeddings $E^C$ for vocabulary $V$ consist of a set of $M$ codebooks $E_1^C$, $E_2^C$, ..., $E_M^C$, each with $K$ basis vectors of the same dimensionality $D$ as the reference embeddings $E$, and a discrete code vector ($C_w^1$, $C_w^2$, ..., $C_w^M$) for each token $w$ in the vocabulary. The final embedding for $w$ is composed by summing up the $C_w^i$th vector from the ith codebook as $E^C(C_w)=\sum_{i=1}^M E_i^C(C_w^i)$. Codebooks and discrete codes are jointly learned using the mean squared distance objective: $(C^*, E^{C*})=\argmin_{C, E^C} \frac{1}{|V|}\sum_{w\in V}||E^C(C_w)-E(w)||^2.$ For learning compositional codes, the Gumbel-softmax reparameterization trick \cite{jang2016categorical, maddison2016concrete} is used for one-hot vectors corresponding to each discrete code.

\begin{table*}[t]
\centering
\resizebox{\textwidth}{!}{%
\begin{tabular}{lcccccccc}
\hline \textbf{Encoder} & \textbf{EncoderParam\# / Size} &\textbf{EmbParam\# / Size} & \textbf{SizeRatio}& \textbf{CCEmbSize} & \textbf{CCEncoderSize} & \textbf{EmbComp} & \textbf{EncoderComp}\\ \hline
RoBERTa-base            & 125.29M / 477.94MB  & 38.60M / 147.25MB   & 30.81\%   & 2.27MB    & 332.96MB & 98.46\%   & 30.33\% \\
BERT-base-uncased	    & 110.10M / 420.00MB  & 23.44M / 89.42MB    & 21.29\%   & 1.97MB    & 332.55MB & 97.80\%   & 20.82\% \\
DistilBERT-base-uncased	& 66.99M / 255.55MB   & 23.44M / 89.42MB    & 34.99\%   & 1.97MB    & 168.10MB & 97.80\%   & 34.22\% \\
ALBERT-large-v2         & 17.85M / 68.09MB    & 3.84M / 14.65MB     & 21.52\%   & 0.71MB    & 54.15MB  & 95.15\%   & 20.47\% \\
ALBERT-base-v2          & 11.81M / 45.05MB    & 3.84M / 14.65MB     & 32.52\%   & 0.71MB    & 31.11MB  & 95.15\%   & 30.94\% \\
\hline
\end{tabular}
}
\caption{\label{adv} Model compression with compositional code (``cc") embeddings. The embedding layers are compressed by more than 95\% with compositional code embeddings in all of the BERT variants.}
\end{table*}

\subsection{Transformer-Based Models with Compositional Code Embeddings}
In this work, we learn compositional code embeddings to reduce the size of the embeddings in pretrained contextualized language models. We extract the embedding tables from pretrained RoBERTa-base \cite{liu2019roberta}, BERT-base \cite{bert}, DistilBERT-base \cite{distilbert}, ALBERT-large-v2 and ALBERT-base-v2 \cite{albert} from the huggingface transformers library \cite{Wolf2019HuggingFacesTS} and follow the approach presented by \citet{codebooks} to learn the code embeddings. We then replace the embedding tables in the transformer models with the compositional code approximations and evaluate the compressed language models by finetuning on downstream tasks. When \citet{codebooks} feed compositional code embeddings into the LSTM neural machine translation model, they fix the embedding parameters and train the rest of the model from random initial values. In our experiments, we fix the discrete codes, initialize the transformer layers with those from the pretrained language models, initialize the task-specific output layers randomly, and finetune the codebook basis vectors with the rest of the non-discrete parameters.

\subsection{Size Advantage of Compositional Code Embeddings}
An embedding matrix $E\in \mathbb{R}^{|V|\times D}$ stored as 32-bit float point numbers, where $|V|$ is the vocabulary size and $D$ is the embedding dimension, requires $32|V|D$ bits. Its compositional code reconstruction requires $32MKD$ bits for $MK$ basis vectors, and $M\log_2K$ bits for codes of each of $|V|$ tokens. Since each discrete code takes an integer value in $[1, K]$, it can be represented using $\log_2K$ bits. 

Table \ref{adv} illustrates the size advantage of compositional code embeddings for various pretrained transformer models \cite{Wolf2019HuggingFacesTS} used in our experiments. While the technique focuses on compressing the embedding table, it is compatible with other compression techniques for transformer models, including parameter sharing among transformer layers and embedding factorization used in ALBERT and distillation for learning DistilBERT. In our experiments, we apply the code learning technique to compress embeddings in five pretrained BERT variants by 95.15\% $\sim$ 98.46\% to build competitive but significantly lighter semantic parsing models.

\begin{table}
\centering
\resizebox{0.45\textwidth}{!}{%
\begin{tabular}{lccccc}
\hline \textbf{Dataset} & \textbf{Train} &\textbf{Valid} & \textbf{Test} & \textbf{\#Intent} & \textbf{\#Slot} \\ \hline
ATIS	& 4,478 & 500 & 893 & 26 & 83 \\
SNIPS	& 13,084 & 700 & 700 & 7 & 39 \\
Facebook TOP    & 31,279 & 4,462 & 9,042 & 25 & 36 \\
\hline
\end{tabular}
}
\caption{\label{table-data} Statistics for semantic parsing datasets.}
\end{table}

\section{Datasets}
Following \citet{rongali}, we evaluate our models on SNIPS \cite{coucke2018snips}, Airline Travel Information
System (ATIS) \cite{price1990evaluation}, and Facebook TOP \cite{gupta2018semantic} datasets for task-oriented semantic parsing (Table \ref{table-data}). For SNIPS and ATIS, we use the same train/validation/test split as \citet{goo2018slot}.

\section{Experiments and Analyses}
For transformer model training, we base our implementation on the huggingface transformers library v2.6.0 \cite{Wolf2019HuggingFacesTS}. We use the AdamW optimizer \cite{loshchilov2017decoupled} with 10\% warmup steps and linear learning rate decay to 0. Forr code embedding learning, we base our implementation on that of \citet{codebooks}. By default we learn code embeddings with 32 codebooks and 16 basis vectors per codebook. Unless otherwise specified, hyperparameters are found according to validation performances from one random run. We conduct our experiments on a mixture of Tesla M40, TITAN X, 1080 Ti, and 2080 Ti GPUs. We use exact match (EM) and intent accuracy as evaluation metrics. Exact match requires correct predictions for all intents and slots in a query, and is our primary metric.

\begin{table}[t]
\centering
\resizebox{0.5\textwidth}{!}{%
\begin{tabular}{lcccccc}
\hline \multicolumn{5}{l}{\textbf{Model}} & \textbf{EM} &\textbf{Intent}  \\ \hline
\multicolumn{5}{l}{Joint BiRNN \cite{hakkani2016multi}}	& 73.2 & 96.9 \\
\multicolumn{5}{l}{Attention BiRNN \cite{liu2016attention}}	&  74.1 & 96.7 \\
\multicolumn{5}{l}{Slot Gated Full Attention \cite{goo2018slot}}	& 75.5 & 97.0 \\
\multicolumn{5}{l}{CapsuleNLU \cite{zhang2019}}	&  80.9 & 97.3 \\
\multicolumn{5}{l}{BERT-Seq2Seq-Ptr \cite{rongali}}	& 86.3 & 98.3 \\
\multicolumn{5}{l}{RoBERTa-Seq2Seq-Ptr \cite{rongali}}	& 87.1 & 98.0 \\
\multicolumn{5}{l}{BERT-Joint \cite{castellucci2019multi}}	& 91.6 & \textbf{99.0} \\
\multicolumn{5}{l}{Joint BERT \cite{chen2019bert}}	& \textbf{92.8} & 98.6  \\
\hline \textbf{Ours} & \textbf{epo} & \textbf{lr} & \textbf{wd} & \textbf{EM-v} &\textbf{EM} & \textbf{Intent} \\ \hline
ALBERT-base	&   & 5e-5 & 0.05 & 90.71 & \textbf{91.29} & 98.86 \\
ALBERT-base\_cc & 1100 & 5e-5 & 0.01 & 90.00 & 89.14 & 98.14 \\\hline
ALBERT-large	&   & 3e-5 & 0.05 & 91.29 & \textbf{92.43} & 98.14 \\
ALBERT-large\_cc & 1100 & 2e-5 & 0.05 & 91.14 & \textbf{92.43} & 98.71 \\\hline
DistilBERT-base	&   & 3e-5 & 0.05 & 90.29 & 91.14 & 98.57 \\
DistilBERT-base\_cc	& 900 & 6e-5 & 0.01 & 90.14 & \textbf{91.24} & 98.43 \\\hline
BERT-base	&  & 3e-5 & 0.05 & 92.14 & \textbf{92.29} & 99.14  \\
BERT-base\_cc	& 900 & 6e-5  & 0.05 & 91.29 & 90.71 & 98.71 \\
\hline
\end{tabular}
}
\caption{\label{table-SNIPS} Results on SNIPS. ``cc" indicate models with code embeddings. ``epo" is the epoch number for offline code embedding learning. ``lr" and ``wd" are the peak learning rate and weight decay for whole model finetuning. ``EM-v", ``EM", ``Intent" indicate validation exact match, test exact match, and test intent accuracy.}
\end{table}

\subsection{SNIPS and ATIS}
We implement a joint sequence-level and token-level classification layer for pretrained transformer models. The intent probabilities are predicted as $\mathrm{y^i=softmax(W^ih_0+b^i)}$, where $\mathrm{h_0}$ is the hidden state of the [CLS] token. The slot probabilities for each token j are predicted as $\mathrm{y^s_j=softmax(W^sh_j+b^s)}$. We use the cross entropy loss to maximize $\mathrm{p(y^i|x)\prod p(y^s_j|x)}$ where j is the first piece-wise token for each word in the query. We learn code embeddings for \{500, 700, 900, 1100, 1300\} epochs. We train transformer models with original and code embeddings all for 40 epochs with batch size 16 and sequence length 128. Uncased BERT and DistilBERT perform better than the cased versions. We experiment with peak learning rate \{2e-5, 3e-5, ..., 6e-5\} and weight decay \{0.01, 0.05, 0.1\}. As shown in Table \ref{table-SNIPS} and \ref{table-ATIS}, we use different transformer encoders to establish strong baselines which achieve EM values that are within 1.5\% of the state-of-the-art.

On both datasets, models based on our compressed ALBERT-large-v2 encoder (54MB) perserves $>$99.6\% EM of the previous state-of-the-art model \cite{chen2019bert} which uses a BERT encoder (420MB). In all settings, our compressed encoders perserve $>$97.5\% EM of the uncompressed counterparts under the same training settings. We show that our technique is effective on a variety of pretrained transformer encoders.

\begin{table}[t]
\centering
\resizebox{0.5\textwidth}{!}{%
\begin{tabular}{lcccccc}
\hline \multicolumn{5}{l}{\textbf{Model}} & \textbf{EM} &\textbf{Intent}  \\ \hline
\multicolumn{5}{l}{Joint-BiRNN \cite{hakkani2016multi}}	& 80.7 & 92.6\\
\multicolumn{5}{l}{Attention-BiRNN \cite{liu2016attention}}	& 78.9 & 91.1\\
\multicolumn{5}{l}{Slot-Gated \cite{goo2018slot}}	& 82.2 & 93.6\\
\multicolumn{5}{l}{CapsuleNLU \cite{zhang2019}}	& 83.4 & 95.0\\
\multicolumn{5}{l}{BERT-Seq2Seq-Ptr \cite{rongali}}	& 86.4 & 97.4 \\
\multicolumn{5}{l}{RoBERTa-Seq2Seq-Ptr \cite{rongali}}	& 87.1 & 97.4\\
\multicolumn{5}{l}{BERT-Joint \cite{castellucci2019multi}}	& \textbf{88.2} & \textbf{97.8} \\
\multicolumn{5}{l}{Joint-BERT \cite{chen2019bert}}	& \textbf{88.2} & 97.5 \\ 
\hline \textbf{Ours} & \textbf{epo} & \textbf{lr} & \textbf{wd} & \textbf{EM-v} &\textbf{EM} & \textbf{Intent} \\ \hline
ALBERT-base	&   & 5e-5 & 0.05 & 93.4 & 86.90 & 97.42 \\
ALBERT-base\_cc & 900 & 6e-5 & 0.1 & 94.2 & \textbf{87.23} & 96.75 \\\hline
ALBERT-large	&   & 5e-5 & 0.05 & 93.8 & \textbf{88.02} & 97.54 \\
ALBERT-large\_cc & 1100 & 5e-5 & 0.1 & 94.0 & 87.91 & 97.54 \\\hline
DistilBERT-base	&   & 4e-5 & 0.05 & 93.6 & \textbf{88.13} & 97.42 \\
DistilBERT-base\_cc	& 1100 & 6e-5 & 0.05 & 93.2 & 87.12 & 97.54 \\\hline
BERT-base	&   &  4e-5 & 0.01 & 93.4 & \textbf{88.13} & 97.54 \\
BERT-base\_cc	& 700 & 6e-5  & 0.1 & 93.0 & 87.35 & 97.20 \\
\hline
\end{tabular}
}
\caption{\label{table-ATIS} Results on ATIS. Refer to the caption of Table \ref{table-SNIPS} for abbreviation explanations.}
\end{table}

\subsection{Facebook TOP}
\begin{table}[t]
\centering
\resizebox{0.5\textwidth}{!}{%
\begin{tabular}{lccc}
\hline \multicolumn{2}{l}{\textbf{Model}} & \textbf{EM} &\textbf{Intent}  \\ \hline
\multicolumn{2}{l}{RNNG \cite{gupta2018semantic}} & 78.51 & - \\ 
\multicolumn{2}{l}{Shift Reduce (SR) Parser} & 80.86 & - \\
\multicolumn{2}{l}{SR with ELMo embeddings} &83.93 & - \\
\multicolumn{2}{l}{SR ensemble + ELMo + SVMRank} & \textbf{87.25} & - \\
\multicolumn{2}{l}{BERT-Seq2Seq-Ptr \cite{rongali}}	& 83.13 & 97.91 \\
\multicolumn{2}{l}{RoBERTa-Seq2Seq-Ptr \cite{rongali}}	& 86.67 & \textbf{98.13} \\
\hline \textbf{Ours} & \textbf{EM-v} &\textbf{EM} & \textbf{Intent} \\ \hline
ALBERT-Seq2Seq-Ptr & 84.56 & \textbf{85.41} & 98.47 \\
ALBERT-Seq2Seq-Ptr\_cc & 83.48 & 84.42 & 98.05 \\ \hline
DistilBERT-Seq2Seq-Ptr & 84.25 & \textbf{85.12} & 98.50 \\
DistilBERT-Seq2Seq-Ptr\_cc & 82.76 & 83.42 & 98.09 \\ \hline
BERT-Seq2Seq-Ptr & 83.83 & \textbf{85.01} & 98.59 \\
BERT-Seq2Seq-Ptr\_cc & 82.36 & 83.34 & 98.25 \\ \hline
RoBERTa-Seq2Seq-Ptr	 & 85.00 & \textbf{85.67} & 98.59 \\
RoBERTa-Seq2Seq-Ptr\_cc	& 83.51 & 83.78 & 98.17 \\ \hline
\end{tabular}
}
\caption{\label{table-BaselineTOP} Results on Facebook TOP. The SR models are by \citet{einolghozati2019improving}. Refer to the caption of Table \ref{table-SNIPS} for abbreviation explanations.}
\end{table}

Table \ref{table-BaselineTOP} presents results on Facebook TOP. We follow \citet{rongali} and experiment with Seq2Seq models. We use different pretrained BERT-variants as the encoder, transformer decoder layers with $d_{model}=768$ \cite{attention}, and a pointer generator network \cite{vinyals2015pointer} which uses scaled dot-product attention to score tokens. The model is trained using the cross-entropy loss with label smoothing of 0.1. For simplicity, we always train code embeddings for 900 epochs offline. Learning rate 2e-5 and weight decay 0.01 are used for transformer training. BERT and DistilBERT are cased in these experiments. During inference, we employ beam decoding with width 5. Our greatly compressed models present 98$\sim$99\% performances of the original models.

\begin{table*}
\centering
\tiny
\resizebox{.75\textwidth}{!}{%
\begin{tabular}{ccccccc}
\hline \textbf{Epoch} & \textbf{MeanEucDist} &\textbf{NN-cos} & \textbf{NN-Euc} & \textbf{SNIPS} & \textbf{ATIS} & \textbf{TOP}\\ \hline
100	    & 0.3677$\pm$0.25\%  & 0.66$\pm$1.90\%        & 0.65$\pm$2.00\% & 79.29 & 82.31 & 78.09 \\
200	    & 0.3254$\pm$0.08\%  & 2.20$\pm$0.69\%         & 2.30$\pm$0.84\% & 85.43 & 84.99 & 81.59 \\
300	    & 0.3023$\pm$0.09\%  & 3.66$\pm$0.92\%         & 3.96$\pm$0.55\%         & 86.86 & 86.11 & 83.17 \\
400	    & 0.2841$\pm$0.23\%  & 4.84$\pm$0.58\%         & 5.26$\pm$0.83\%         & \textbf{89.71} & 87.01 & 83.45 \\
500	    & 0.2685$\pm$0.26\%  & 5.72$\pm$0.48\%         & 6.21$\pm$0.78\%         & 87.71 & 87.23 & 83.82 \\
600	    & 0.2573$\pm$0.12\%  & 6.20$\pm$0.39\%         & 6.72$\pm$0.18\%         & 88.14 & 85.69 & 83.41 \\
700	    & 0.2499$\pm$0.20\%  & 6.42$\pm$0.49\%         & 6.94$\pm$0.33\%         & 88.00 & \textbf{87.35} & 84.27 \\
800	    & 0.2444$\pm$0.07\%  & 6.54$\pm$0.39\%         & 7.07$\pm$0.15\%         & 88.57 & 86.90 & 84.09 \\
900	    & 0.2407$\pm$0.10\%  & 6.62$\pm$0.31\%         & 7.14$\pm$0.14\%         & 88.57 & 86.56 & \textbf{84.42} \\
1000	& 0.2380$\pm$0.07\%  & 6.65$\pm$0.39\%         & 7.16$\pm$0.10\%         & 89.14 & 87.12 & 83.86 \\
\hline
\end{tabular}
}
\caption{\label{table-emp} Analyses for the code embedding learning process (M=32, K=16). MeanEucDist, NN-cos, and NN-Euc are averaged across 5 runs. ``SNIPS", ``ATIS", and ``TOP" are the test exact match achieved on the three datasets.}
\end{table*}

\subsection{Analysis for Code Convergence}
We study the relationship among a few variables during code learning for the embeddings from pretrained ALBERT-base (Table \ref{table-emp}). During the first 1000 epochs, the mean Euclidean distance between the original and reconstructed embeddings decrease with a decreasing rate. The average number of shared top-20 nearest neighbours according to cosine similarity and Euclidean distances between the two embeddings increase with a decreasing rate. We apply code embeddings trained for different numbers of epochs to ALBERT-base-v2 and finetune on semantic parsing. On SNIPS and ATIS, we find the best validation setting among learning rate \{2,3,4,5,6\}e-5 and weight decay \{0.01, 0.05, 0.01\}. We observe that the test exact match plateaus for code embeddings trained for more than 400 epochs. On Facebook TOP, we use learning rate 2e-5 and weight decay 0.01, and observe the similar trend.

\subsection{Effects of M and K}
We use embeddings from pretrained ALBERT-base-v2 as reference to learn code embeddings with M in \{8, 16, 32, 64\} and K in \{16, 32, 64\}. As shown in Table \ref{table-mk}, after 700 epochs, the MSE loss for embeddings with \textit{larger} M and K converges to \textit{smaller} values in general. With M=64, more epochs are needed for convergence to smaller MSE losses compared to those from smaller M. We apply the embeddings to ALBERT-base-v2 and finetune on SNIPS. In general, larger M yields better performances. Effects of K are less clear when M is large.

\begin{table}[t]
\centering
\tiny
\resizebox{.4\textwidth}{!}{%
\begin{tabular}{cccccc}
\hline \textbf{M} & \textbf{K} & \textbf{epo} & \textbf{MSE} & \textbf{EM}\\ \hline
8 & 16 & 700 & 0.3155$\pm$0.05\% & 85.43 \\
8 & 32 & 700 & 0.3032$\pm$0.04\% & 87.43 \\
8 & 64 & 700 & 0.2944$\pm$0.04\% & 87.43\\
16 & 16 & 700 & 0.2855$\pm$0.05\% & 88.57\\
16 & 32 & 700 & 0.2727$\pm$0.09\% & 88.00\\
16 & 32 & 700 & 0.2669$\pm$0.08\% & 88.14\\
32 & 16 & 700 & 0.2499$\pm$0.20\% & 89.00\\
32 & 32 & 700 & 0.2421$\pm$0.20\% & 89.14\\
32 & 64 & 700 & 0.2396$\pm$0.27\% & 88.29\\
64 & 16 & 700 & 0.2543$\pm$0.47\% & 88.29\\
64 & 16 & 1000 & 0.2256$\pm$1.06\% & 89.71\\
64 & 32 & 700 & 0.2557$\pm$0.37\% & 89.86\\
64 & 32 & 1000 & 0.2159$\pm$0.43\% & 89.71\\
\hline
\end{tabular}
}
\caption{\label{table-mk} Effects of M and K. Mean squared errors (MSE) are averaged over 5 runs. Best validation exact match (EM) is presented for compressed transformer models trained with 0.05 weight decay and \{3,4,5,6,7\}e-5 peak learning rates on SNIPS.}
\end{table}

\section{Conclusion}
Current state-of-the-art task-oriented semantic parsing models are based on pretrained RoBERTa-base (478MB) or BERT-base (420MB). We apply DistilBERT (256MB), ALBERT-large (68MB), and ALBERT-base (45MB), and observe near state-of-the-art performances. We learn compositional code embeddings to compress the model embeddings by 95.15\% $\sim$ 98.46\%, the pretrained encoders by 20.47\% $\sim$ 34.22\%, and observe 97.5\% performance preservation on SNIPS, ATIS, and Facebook TOP. Our compressed ALBERT-large is 54MB and can achieve 99.6\% performances of the previous state-of-the-art models on SNIPS and ATIS. Our technique has potential to be applied to more tasks including machine translation in the future.

\section*{Acknowledgement}
This project is part of the data science industry mentorship program initiated by Andrew McCallum at University of Massachusetts Amherst. We thank the teaching assistants Rajarshi Das and Xiang Lorraine Li for helpful discussion and the instructor Andrew McCallum for valuable feedback. Experiments in this project are conducted on the Gypsum cluster at UMassAmherst. The cluster is purchased with funds from the Massachusetts Technology Collaborative.

\bibliographystyle{acl_natbib}
\bibliography{anthology,emnlp2020}

\end{document}